\pdfoutput=1
\documentclass[11pt]{article}
\usepackage{EACL2023}
\usepackage{times,todonotes}
\usepackage{latexsym}

\usepackage{microtype}
\usepackage{amsmath}

\title{Scalable Prompt Generation for Semi-supervised Learning with Language Models}

\author{Yuhang Zhou* \\
 University of Maryland \\
 College Park, MD \\
  \texttt{tonyzhou@umd.edu} \\\And
  Suraj Maharjan* \\
 Amazon\\
 Seattle, WA\\
  \texttt{mhjsuraj@amazon.com} \\ \And
  Beiye Liu \\
  Amazon \\
   New York, NY\\
  \texttt{beiyeliu@amazon.com} \\}

\date{}

\begin{document}
\maketitle
\begin{abstract}
Prompt-based learning methods in semi-supervised learning (SSL) settings have been shown to be effective on multiple natural language understanding (NLU) datasets and tasks in the literature. However, manually designing multiple prompts and verbalizers requires domain knowledge and human effort, making it difficult and expensive to scale across different datasets. In this paper, we propose two methods to automatically design multiple prompts and integrate automatic verbalizer in SSL settings without sacrificing performance. The first method uses various demonstration examples with learnable continuous prompt tokens to create diverse prompt models. The second method uses a varying number of soft prompt tokens to encourage language models to learn different prompts. For the verbalizer, we use the prototypical verbalizer to replace the manual one. In summary, we obtained the best average accuracy of 73.2\% (a relative improvement of 2.52\% over even the previous state-of-the-art SSL method with manual prompts and verbalizers) in different few-shot learning settings.\let\thefootnote\relax\footnote{*Equal contribution. This work was done during Yuhang's internship at Amazon, Alexa AI.}
	
\setcounter{footnote}{0}
% \todo[author=Ben]{This is not intuitive. We need to show relative comparison to SOTA solution}
%relative improve = (71.5-70.8)/70.8*100 = 0.98870056497

%relaitve improve = (73.2-71.4)/71.4 *100 =2.521008403
\end{abstract}

% The paper tackles the problem with designing manual prompts and verablizers and proposes novel methods to automate their design process in semi-supervised learning settings. This makes it easy to scale across multiple tasks and datasets. Moreover, the research might open new door toward prompt based few-shot learning era.

% not to pubish
%  This algorithm might be used in future for 3P IC/NER tasks.

\section{Introduction}

Pre-training large language models with huge amounts of text corpora in masked language modeling tasks and then fine-tuning the pre-trained language model (PLM) on downstream tasks have shown superior performance in many natural language processing tasks. However, the discrepancy between the pretraining task (masked language modeling objective) and the downstream fine-tuning task (task without MASK token) could lead to unexpected behaviors. Recently, there has been growing research interest in the area of prompt-tuning, where any NLU task is transformed into a cloze task to mimic the pre-training objective of a large masked language model \cite{kumar2016ask, mccann2018natural, radford2018improving}. Prompt-based learning transforms an input $\mathbf{x}$ into $\mathbf{x}'$ using a prompt function. It makes use of the vast amount of acquired knowledge of PLMs to predict a distribution of tokens at the masked position. The verbalizer then maps the predicted tokens to classes. The main advantage of this approach is that this method works well in a few-shot learning environment \cite{schick-schutze-2021-exploiting}. However, the main disadvantage of this method is the limitation posed by the prompt and verbalizer functions, which require human knowledge to carefully craft them. Such handcrafting work is expensive and not scalable with the increase in the variety of tasks and datasets. For example, in Alexa, there are thousands of domains and manually designing prompts and verbalizer for intent classification for each of them according to the dataset content demand human expertise, which is time consuming and not applicable. It is essential to reduce the human efforts in the process of prompt generation. Prompt-based learning requires finding the right tokens in the prompts that align with the task requirement and dataset content. However, since the objective of these prompt tokens is only for the language models to perform the task at hand, it is not necessary for them to be a sequence of words that humans can understand. 

Continuous prompt-based learning alleviates the need for human intervention to determine prompt tokens. Instead, it automates the prompt design process. In the literature, there are mainly two methods: i) automatically search for discrete prompt text tokens~\cite{shin-etal-2020-autoprompt} ii) automatically learn numerical prompt embeddings~\cite{lester-etal-2021-power,li-liang-2021-prefix,liu2021gpt,P2,hambardzumyan-etal-2021-warp}. The main difference between these two approaches is that the first searches for actual discrete tokens from the language model vocabulary, whereas the second method directly learns the embeddings for prompt tokens, which may not be human comprehensible. Similarly, automatic selection of label words~\cite{shin-etal-2020-autoprompt,schick-etal-2020-automatically,gao-etal-2021-making}, soft verbalizer~\cite{hambardzumyan-etal-2021-warp,P2}, and prototypical verbalizer~\cite{cui-etal-2022-prototypical} are the methods proposed to eliminate the tedious process of manually defining verbalizer mapping functions.

Most of these continuous prompt and automatic verbalizer methods focus on supervised learning (SL) settings but ignore their generalization under semi-supervised learning (SSL) settings. The previous state-of-the-art (SoTA) SSL method with various manual prompts and verbalizers has shown superiority over SL language models with a single manual prompt \cite{schick-schutze-2021-exploiting}. In this SSL pipeline, we normally train several labeler models with different manual prompts to capture diverse information from the limited training data and make use of them to annotate a huge amount of unlabeled data. Having to design several manual prompts and verbalizer models for SSL settings and applying them across multiple datasets and tasks will exacerbate the scalability and cost problem. In this paper, we tackle the problem posed by manual prompt and verbalizer design and propose automatic methods to fully automate the design of diverse prompts and verbalizers in SSL settings. Our main contributions are as follows.
\begin{itemize}

%     \item We empeliminate the use of manual verbalizers, which requires human involvement and thus expensive, with prototypical verbalizers in SSL settings. We also empirically show that the prototypical verbalizers are comparable with manual verbalizers through experimentation on multiple datasets on different NLP tasks.
    
%     \item We propose a method to generate diverse prompts by adding multiple demonstration examples with continuous prompt tokens for using in SSL settings. This method ensures that the continuous prompt can be applied inside the SSL loop to avoid the human involvement.
    
%   \item To our best knowledge, we are the first to propose a semi-supervised learning method by using the continuous prompts and automatic verbalizers that provide state-of-the-art results while removing all per-dataset manual engineering.
    \item We propose methods to generate various prompts by adding multiple demonstration examples with continuous prompt tokens for use in SSL settings. 
   \item To the best of our knowledge, we are the first to completely eliminate human involvement in designing multiple prompts and verbalizers in SSL settings and obtain similar and even better performance than the SoTA methods with manual prompts and verbalizers.
    \item We empirically show that using the automatic verbalizer with manual prompts can achieve a similar performance to manual verbalizers' performance in the SSL pipeline.
   
%   \item We demonstrate the success of our proposed methodology on both academic and real-world industry benchmarks on scale.

\end{itemize}
\section{Methodology}
\begin{figure*}[!htb]
  \centering
  \includegraphics[width=\textwidth] 
  {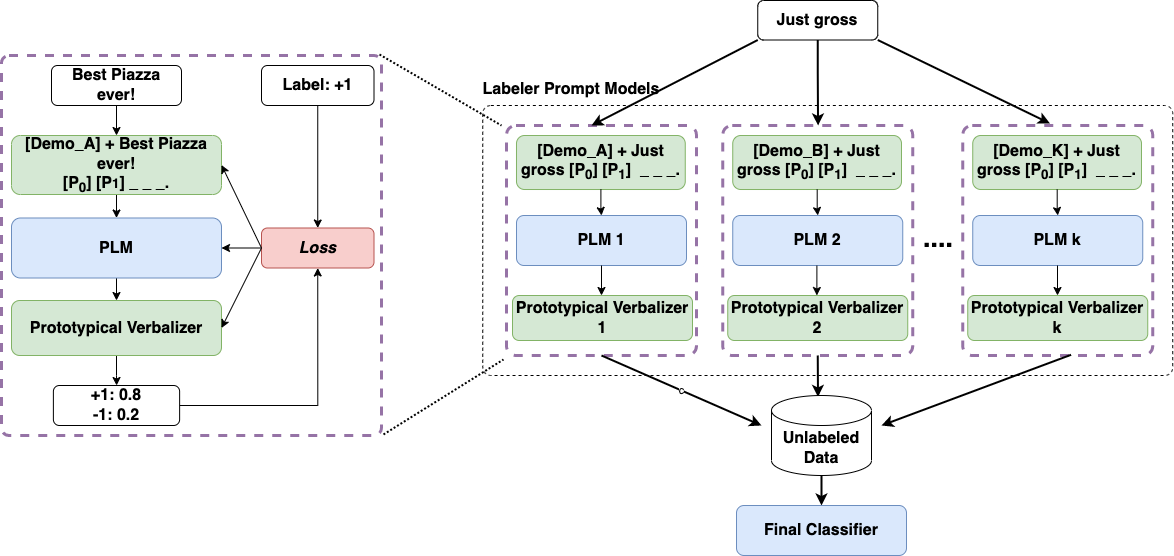}
   \caption{\label{fig:ssl} Semi-Supervised Learning (SSL) Training. Multiple diverse prompt-based learning models are trained on labeled data to soft label huge amounts of unlabeled data. The soft labels serve as ground truth to train the final classifier. $P_0,P_1,\ldots$ are continuous prompt tokens and $Demo\_A,Demo\_B,\ldots$ are demonstration examples randomly sampled from the training data.}
\end{figure*}

Our overall prompt-based SSL workflow follows Pattern-exploiting Training (PET) semi-supervised learning setting \cite{schick-schutze-2021-exploiting}. PET first transforms the input sequence $x$ to a cloze question containing a single MASK token. Next, it uses PLM to fill in the value of the MASK token and applies verbalizers to map the output tokens to the class labels $y \in Y$. They devise a semi-supervised framework to produce soft labels on a large amount of unlabeled data, which are later used to train a final supervised classifier $\mathbf{F}$. They report strong performance over other supervised prompt-tuning methods and other semi-supervised approaches without prompts across multiple NLU tasks. Before this paper, the PET approach was the state-of-the-art (SoTA) framework that integrates the prompt-tuning method into the SSL pipeline.

The PET method fine-tunes multiple PLMs with different prompts. It introduces diversity in the prompts by manually designing several prompts using domain and task knowledge. Similarly, it uses human expertise to design verbalizer mappings for each of the datasets based on the knowledge of the tasks. Here, we use continuous and automatic prompts and verbalizers, thus eliminating the need for human involvement in designing manual prompts and verbalizers.

\subsection{Overall Pipeline}
Figure~\ref{fig:ssl} shows the overall pipeline of our proposed methods. Unlike the original PET pipeline with manual prompts and verbalizers, we use a prompt generation function to generate multiple automatic prompts. Each PLM with automatic prompts serves as a labeler model. We train each of these prompts $+$ automatic verbalizer models with a labeled dataset $\mathcal{T}$ in few-shot settings. With an input sequence $x_t \in \mathcal{T}$ and the given label $y_t$, we first use the prompt function $P$ to transform $x_t$ into a sequence $P(x_t)$ with a MASK token. The verbalizer then maps the predicted word probability at the masked position to the label probability. For each PLM $m$, the predicted probability $p_m (y_t | x_t)$ is defined as
\begin{equation}
    p_m (y_t | x_t) = \frac{\exp m(y_t | x_t)} {\sum_{y' \in Y} \exp m(y' | x_t)}
\end{equation}
where $m(y | x)$ is the raw score of PLM $m$ in the masked position. After obtaining the probability, we minimize the cross-entropy loss $\mathcal{L}_c$ between $p_m (y | x)$ and $y$.

We apply trained labeler models to each sentence $x_d \in \mathcal{D}$ in the unlabeled dataset $\mathcal{D}$ and get the probability $p_m (y_d | x_d)$ for each trained model. We then take the average of these probabilities from each trained model $m$ as the ground-truth probability,
\begin{equation*}
    p_t(y_d | x_d) = \frac{1}{Z} \sum_{m \in M} p_m (y_d | x_d)
\end{equation*}
where $Z$ is the total number of trained PLMs with different automatic prompts.
Eventually, we fine-tune a final pre-trained language model $\mathbf{F}$ with a standard sequence classification head. We use the Kullback-Leibler (KL) divergence as our loss function. Given $p_t(y_d | x_d)$ and the predicted probability $\hat{p} (y_d | x_d)$ of the final classifier $\mathbf{F}$, the divergence loss $\mathcal{L}_{div}$ for this input is:
\begin{equation}
    \mathcal{L}_{div} (x_d) = \sum_{y' \in Y} p_t (y' | x_d) \log \left(\frac{p_t (y' | x_d)}{\hat{p} (y' | x_d)}\right)
\end{equation}
The final classifier $\mathbf{F}$ is then applied to the test set to obtain the results.

\citet{schick-schutze-2021-exploiting} introduce diversity in their SSL pipeline by training several models with different manual prompts and applying them to softly label a large number of unlabeled datasets. The diversity between manual prompts brings consistent improvements. We observe that diverse knowledge learned by the language model is mostly introduced by the prompts rather than manual verbalizers, since in most datasets, they prepare only one manual verbalizer but multiple prompts for experimentation. Thus, we propose replacing manual prompts with multiple automatic prompts and using the same automatic verbalizer for all labeler models.

% Due to the absence of a large development dataset in the few-shot setting, it is hard to identify which prompts and verbalizer performed well. Their strategy was to ask all trained language models to annotate the unlabeled dataset and aggregate the soft labels. Their strategy ensured that the final classifier $\mathbf{F}$ can learn the diverse knowledge from language models trained with different prompts.

\subsection{Continuous Prompt Design}
Several researchers have proposed methods to automate the prompt design process \cite{liu2021gpt, li-liang-2021-prefix, lester-etal-2021-power}. In most of these methods, they insert the continuous trainable prompt tokens into the input sentence and learn the token embeddings during the training process. However, existing continuous prompt-based learning methods do not consider their application in the PET pipeline, which requires training several labeler models \cite{schick-schutze-2021-exploiting}, in order to learn diverse knowledge from the datasets. Therefore, most methods do not define strategies to compose multiple continuous prompts. We propose two scalable solutions to introduce different variables in the design of continuous prompt labeler models (various demonstration examples or varying numbers of continuous prompt tokens). We expect that with these diverse continuous prompts, trained language models can fully learn different aspects of knowledge from the training dataset. 

% to induce the PLMs to compose different prompts.
% which means when given a single few-shot dataset, the existing methods do not provide a strategy to compose multiple prompts.

\subsubsection{Scalable Prompt Generation}
Inspired by the P-tuning \cite{liu2021gpt} method, we insert multiple continuous prompt tokens $p_n$ into the input sentence $x$, transforming it into $[\mathbf{x}][p_0,p_1,\ldots,p_n][\text{MASK}].$. Different from the original P-tuning method, we invent two scalable designs to make it suitable for the prompt-based SSL pipeline.

\label{sec:add_demo}
\paragraph{Add Demonstration Examples:}
In this method, we add different demonstration examples to construct diverse prompts. This is similar to the prompt augmentation method, in which one chooses to add additional answered prompts to demonstrate what kind of answer the language model should produce for the MASK token \cite{liu2021pre}. These additional answered prompts are called the demonstration example $[demo]$. To reduce the discrepancy between the demonstration examples and the input sentences, we also add a fixed number of continuous prompt tokens $p$ between the demonstration sentence and its true label. Thus, given the labeled input $\mathbf{x_d}$ and its corresponding ground-truth label $\mathbf{y_d}$ from the labeled training dataset, we construct the demonstration example as $[demo] = [\mathbf{x_d}][p_0,p_1,\ldots,p_n][\mathbf{y_d}]$, where $p_0,p_1,\ldots,p_n$ are continuous prompt tokens.

After composing the demonstration examples $[demo]$, given a training input from the labeled dataset $x_t=(s_i,s_2,\ldots,s_k) \in \mathcal{T}$ and label $y_t$, where $s_i,s_2,\ldots,s_k$ are input tokens for the PLM $m$, the prompt template function $P_1(x_t)$ is formally defined as
\begin{equation}
\label{eq:demo}
\begin{split}
    & P_1(x_t)_1 = [demo_1][\mathbf{x_t}][p_0,\ldots,p_n][\text{MASK}] \\
    & \ldots \\
    & P_1(x_t)_k = [demo_k][\mathbf{x_t}][p_0,\ldots,p_n][\text{MASK}]
\end{split}
\end{equation}
We create multiple prompts by adding different demonstration examples with exactly $n$ continuous soft tokens with the input sentence. Demonstration examples are randomly sampled from the labeled datasets. For longer input sentences, we first truncate the length of $[demo]$ to fit the PLM requirement. Our intuition is that different demonstration examples will introduce the diversity necessary for SSL experimentation.
%We prepare demonstration examples from different classes, which are randomly sampled from the labeled dataset, for each labeler model. For the long input sentence $x_t$, we first truncate the length of $[demo]$ to fit the PLM requirement. Our intuition is that multiple demonstration examples with different ground-truth labels will introduce the diversity necessary for SSL experimentation. For example, if five PLMs are required as labeler models, we prepare five different demonstration examples from Class 1 to Class 5 to train the PLM, respectively. \footnote{If the total number of class is less than 5, we will sample multiple examples from each class.} We expect that each labeler model fully learns the knowledge of one class and the final classifier $\mathbf{F}$ can distill the diversity. 

\paragraph{Vary Soft Token Numbers:}
In this method, we vary the number of continuous prompt tokens between different labeler models. In other words, this prompt function $P_2(x_t)$ with input sentence $x_t$ is defined as
\begin{equation}
\label{eq:vary}
\begin{split}
& P_2(x_t)_1 = [\mathbf{x_t}][p_0,p_1,\ldots,p_{n_1}][\text{MASK}] \\
& \ldots \\
& P_2(x_t)_k = [\mathbf{x_t}][p_0,p_1,\ldots,p_{n_k}][\text{MASK}]
\end{split}
\end{equation}
and each of the labeler models uses different $n_1$ to $n_k$ number(s) of continuous prompt tokens $p$. Here, we do not prepend the demonstration example. Our intuition is that given different numbers of continuous prompt tokens, the optimized learned continuous prompts may also be different. For example, for AG's News dataset \cite{zhang2015character} about news topics, the optimized prompts with two continuous prompt tokens could be: $[[\mathbf{x}][\text{News : }][\text{MASK}]]$, while optimized prompts with three continuous prompt tokens could be: $[[\mathbf{x}][\text{the category is}][\text{MASK}]]$. We expect that varying the number of continuous prompt tokens will have a similar impact to manually constructing different prompts.

% The second transforms the input into $[\mathbf{x}][p_0,p_1,\ldots,p_{n'}][\text{MASK}].$ Here, we do not prepend the demonstration example but instead, change the number of continuous tokens across multiple labeler models. %%pcom
% Our intuition is that the different demonstration examples as well as varying number of continuous tokens will introduce the diversity necessary for SSL experimentation.

\subsubsection{Reparameterization Block}
\label{sec:reparameterization}
% \citet{li-liang-2021-prefix} and ~\citet{liu2021gpt} empirically show that directly updating the prompt parameters leads to unstable optimization.~\citet{li-liang-2021-prefix} introduce reparameterization of the prompt embeddings by passing the prompt embeddings first through a feedforward neural network, while ~\citet{liu2021gpt} use a bidirectional LSTM~\cite{hochreiter:1997} network with two-layer $ReLU$ activated multilayer perceptron (MLP) before feeding the prompt embeddings to MLM. Here, we also follow the approach of the reparameterization block~\citep{liu2021gpt} and first feed the prompt vectors $p_i$ through a bidirectional LSTM network. 
\citet{li-liang-2021-prefix} and ~\citet{liu2021gpt} empirically show that directly updating the parameters in continuous prompts leads to unstable optimization. Hence, we first feed prompt embeddings through a reparameterization block rather than directly feeding them into the PLM. 
%Before feeding the initialized soft token embeddings into the PLM, we first pass them into a reparameterization block, which can increase the stability of training performance, introduced by previous studies \cite{liu2021gpt, li-liang-2021-prefix}. 
Our reparametrization block uses a bidirectional LSTM~\cite{hochreiter:1997} network with a two-layer $ReLU$ activated multilayer perceptron (MLP)~\cite{liu2021gpt, li-liang-2021-prefix}.

 We denote the random initialized tokens as $p_i'$ and the real input embeddings, which are fed into the PLM, as $p_i$. The  $p_i$ are the output of the bidirectional LSTM network and the MLP as, 
 %$p_i = \text{MLP}([\text{LSTM}(p_{0:i}'), \text{LSTM}(p_{i:n}')])$,
\begin{equation*}
    p_i = \text{MLP}([\text{LSTM}(p_{0:i}'), \text{LSTM}(p_{i:n}')])
\end{equation*}
where $p_i$ is also the soft token used in Equations \ref{eq:demo} and \ref{eq:vary}.We learn the optimized continuous prompt tokens $\hat{p}_{0:n}$ during the training process. With the downstream cross-entropy loss $\mathcal{L}_c$, we can differentially optimize the continuous prompts by:
\begin{equation}
    \hat{p}_{0:n} = \underset{p}{\mathrm{argmin}} \mathcal{L}_c (p_m(x | y), y)
\end{equation}

% By also updating the parameters in the reparametrization block during the training process, we can find the optimized continuous prompt tokens $\hat{p}_{0:n}$.
% \todo[author=Ben]{ Are we using this?}

% $$
% [demonstration.text_a] + [demonstration.text_b] + 4 [soft tokens] + [demonstration.label] + [sep] + [placeholder.text_a] + [placeholder.text_b] + 4 [soft tokens] + [masked]
% $$
% \todo{- reparametrization as used in Ptuning
% --NLI , Text clf example explanaition
% }

% Finally, we  optimize the prompt tokens $p_i' (0 \leq i \leq n)$ with the cross entropy loss $\mathcal{L}_c$:
% \begin{equation}
%     \hat{p}_{0:n} = \underset{p'}{\mathrm{argmin}} \mathcal{L}_c (p_m(x | y), y)
% \end{equation}

\subsection{Automatic Verbalizers}
\label{sec:verbalizer}
There are several automatic verbalizer methods that eliminate the need for human intervention and expertise to build mapping functions. We experiment with three types of automatic verbalizers: i) soft verbalizer~\cite{hambardzumyan-etal-2021-warp}, ii) prototypical verbalizer~\cite{cui-etal-2022-prototypical}, and iii) search-based verbalizer~\cite{schick2020automatically}.

\citet{cui-etal-2022-prototypical} experimentally show the superiority of the prototypical verbalizer in a supervised learning environment. However, they did not conduct such experiments for SSL settings. Our experiment with the SSL PET method (details in Section \ref{experiment_results}) with different automatic verbalizers showed that the prototypical verbalizer performed better than the soft verbalizer and the search-based verbalizer on multiple datasets. Thus, we choose to use the prototypical verbalizer as a replacement for the manual verbalizer.

With the optimized embedding of the MASK token from PLM $m$ and the ground-truth labels $y$, the prototypical verbalizer learns the prototype vectors for each class using contrastive learning~\cite{oord2018representation}. The prototypical verbalizer first initializes a prototype embedding for each class label and then uses the embedding of the MASK token as the instance embedding. It uses instance-instance loss $\mathcal{L}_{ins}$ to maximize intra-class similarity and minimize inter-class similarity. Similarly, it uses instance-prototype loss $\mathcal{L}_{proto}$ to maximize the similarity between the prototype and instances belonging to the same class and minimize the similarity of instances belonging to other classes. The probability distribution of the MASK token for each class is calculated by the cosine similarity between the instance embedding and each optimized prototype embedding. For inference, it assigns the class of the prototype vector to the instance with the highest probability score, which is computed by taking the similarity scores of the instance vector with the prototype vectors and normalizing them.

\subsection{Training and Inference Strategy}
All model parameters to be optimized are randomly initialized. As mentioned in Section \ref{sec:reparameterization} and \ref{sec:verbalizer}, we update the parameters in the continuous prompts and PLMs with the loss $\mathcal{L}_c$ and optimize the parameters in the verbalizers with the loss $\mathcal{L}_{ins}$ and $\mathcal{L}_{proto}$. Instead of summing all losses together, our training strategy is to first freeze the parameters in the prototypical verbalizer and then train the parameters in the reparameterization block and the PLM together with the cross-entropy loss $\mathcal{L}_{c}$. Then we freeze the learned parameters and train the parameters in the prototypical verbalizers with instance-instance loss $\mathcal{L}_{ins}$
and instance-prototype loss $\mathcal{L}_{proto}$. After training all labeler models and obtaining the class probability on the unlabeled dataset, we use $\mathcal{L}_{div}$ to fine-tune the final language model classifier. During inference, we do not rely on any prompt-based labeler models and directly use the final fine-tuned language model $\mathbf{F}$ to predict on the test dataset.

\section{Experiments}
To verify the effectiveness of our framework, we conduct multiple semi-supervised learning experiments with several strong baseline frameworks on the commonly-used NLU benchmarks.

\subsection{Dataset Collection}

We experiment with five different datasets\footnote{We downloaded these datasets using the script provided by OpenPrompt~\url{https://github.com/thunlp/OpenPrompt}}: AG’s News~\cite{zhang2015character}, Yahoo Answers~\cite{NIPS2015_250cf8b5}, MNLI (MultiNLI, Multi-Genre Natural Language Inference,~\citet{williams-etal-2018-broad}), RTE (Recognizing Textual Entailment,~\citet{rte}) and CB (CommitmentBank,~\citet{cb}). AG’s News and Yahoo answers are topic classification (TC) datasets, while MNLI, RTE, and CB are natural language inference (NLI) datasets. In Table~\ref{table:dataset}, we provide the number of distinct classes, the unlabeled dataset size used for SSL, and the test size for all five datasets. Details about the design of prompts and verbalizers can be found in Appendix \ref{templates}. 

\begin{table}[!htb]
\centering
\resizebox{\columnwidth}{!}{%
\begin{tabular}{|llrrr|}\hline
\textbf{Dataset} & \textbf{Task} & \textbf{\#Class} & \textbf{\#Unlabeled }& \textbf{\#Test} \\\hline\hline
AG’s News    & TC     & 4        & 40,000       & 7,600  \\
Yahoo        & TC     & 10       & 100,000      & 60,000 \\\hline\hline
CB           & NLI    & 3         & 30,000       & 56      \\
RTE          & NLI    & 2         & 20,000       & 277       \\
MNLI         & NLI    & 3         & 30,000       & 9,815      \\\hline
\end{tabular}
}
\caption{\label{table:dataset} Data statistics. TC= Topic Classification, NLI= Natural Language Inference}
\end{table}

We perform multiple experiments in few-shot settings for all datasets. For few-shot experiments, we use $1, 5, 10, 20$ examples per class for all datasets except for CB and RTE, where we experiment with $32$ examples to align with earlier research work~\cite{schick-schutze-2021-exploiting}. We report the average accuracy for the evaluation across three runs of each experiment with three different random seeds.
\subsection{Proposed Models}
\noindent {\bf Demo+Soft Tokens PET}: The first method is to replace the manual verbalizer with the prototypical verbalizer and manual prompts with demonstration examples and continuous prompt tokens.

\noindent {\bf Vary Soft Tokens PET}: The second method is to introduce diversity by varying the number of continuous prompt tokens, and we use the prototypical verbalizer across multiple labeler models.

\subsection{Models for Comparison}
We design several strong baseline experiments in addition to our proposed models and also perform an ablation study to show the superiority of our proposed models in multiple NLU tasks.
% Besides our proposed prompting models, we also devise several strong baseline experiments and the ablation study to show the superiority of our proposed models on multiple NLU tasks.

\subsubsection{Baseline Models}
\noindent {\bf Fine-tune}: This is a supervised method, where we directly fine-tune the RoBERTa-large PLM with training examples in different few-shot settings. In this method, we do not leverage the unlabeled data.

\noindent {\bf Prototypical Verbalizer PET}: This is a semi-supervised learning method similar to ~\citet{schick-schutze-2021-exploiting}, but we replace the manual verbalizer with the prototypical verbalizer and keep the manual prompts. Experiments with this setup will show the benefits of applying automatic verbalizer in the PET framework.

\noindent {\bf Manual PET}: This is a semi-supervised learning method from~\citet{schick-schutze-2021-exploiting}. Our main goal is to show that, with our proposed method, we can achieve similar or better results than this manual method.

There are other SSL methods that rely on data augmentation without prompt tuning, such as UDA \cite{xie2020unsupervised} and MixText \cite{chen2020mixtext}. Since their performance is consistently worse than the Manual PET model across multiple datasets \cite{schick-schutze-2021-exploiting}, we do not choose these models for comparison in this work.
% \noindent{\bf p-tuning w/o demos and fixed soft tokens PET final (SSL)}:
% We
\subsubsection{Model Intervention for Ablation Study}
\noindent {\bf Fixed Soft Tokens PET}: This semi-supervised learning method is similar to our second proposed method, where we vary the number of continuous tokens to create multiple prompts. However, here we keep the number of continuous tokens fixed and do not add demonstration examples as well. This experiment will help us to understand the importance of diversity introduced by varying continuous tokens in prompt design.

\noindent {\bf Demo+Soft in SL}: This is a supervised method, where we use a prompt template to transform the input by adding a randomly selected demonstration example from the training data and a fixed number of continuous prompt tokens to the input, and we use the prototypical verbalizer for classification. We use RoBERTa-large for PLM. With this experiment, we try to understand the power of semi-supervised learning methods with multiple prompts over supervised training.

\subsection{Implementation Details}

We use the RoBERTa-Large model ~\cite{roberta} as our PLM for all of our experiments. We use AdamW as our optimizer with a learning rate of $1\mathrm{e}{-5}$ and a weight decay of $0.01$ with linear scheduler, batch size of $2$, and trained for $5$ epochs. The reparameterization block contains 2-layer bidirectional LSTM and 2 linear layers with ReLU activation function. The hidden dimension of the linear layer and LSTM layer is 768, as well as the hidden dimension of Roberta-Large. We train the parameters in the reparameterization block and the PLM together. For the prototypical verbalizer, we base our implementation on the Pytorch\footnote{\url{https://pytorch.org/}}, Huggingface transformer\footnote{\url{https://huggingface.co/}},  and OpenPrompt\footnote{\url{https://github.com/thunlp/OpenPrompt}} frameworks~\cite{ding2021openprompt}. For our Demo+Soft Tokens PET, each labeler model will learn 5 soft tokens with different demonstrations. For our Vary Soft Tokens PET, we prepare 5 prompts for each dataset and the number of soft tokens in each prompt ranges from 1 to 5.
\subsection{Results of Multiple Automatic Verbalizers}
\label{experiment_results}
\begin{table}[!h]
\centering
\resizebox{\columnwidth}{!}{%
\begin{tabular}{|l|c|rrr|}
\hline
\textbf{Datasets}        &            & \multicolumn{3}{|c|}{\textbf{SSL PET }}                        \\\hline
\textbf{}                & \textbf{\# instances} & \textbf{SoftVerb} & \textbf{SearchVerb} & \textbf{ProtoVerb } \\\hline\hline
AG’s News     & 10      & 49.4        & \textbf{80.5}   & 77.2             \\
Yahoo         & 10      & 11.8        & 34.0            & \textbf{51.9}   \\\hline\hline
CB           &  32      &  {\bf 88.7}           & 73.2            & 85.7     \\
RTE           & 32      & 48.2        & 50.2            & \textbf{52.8}    \\
MNLI          & 10      & 39.0        & 37.0            & \textbf{50.0}  \\\hline        
\end{tabular}
}
\caption{\label{table:verbalizer} Average accuracy on different datasets by replacing manual verbalizers with automatic verbalizers in the PET SSL setup. For CB and RTE, we use 32 training examples, whereas for other datasets, we use 10 training examples to train labeler models. The best performance is marked in bold.}
\end{table}

To understand which automatic verbalizer is a better replacement for manual verbalizer, we first experiment with three automatic verbalizers: soft verbalizer~\cite{hambardzumyan-etal-2021-warp,liu2021gpt,P2}, search verbalizer~\cite{gao-etal-2021-making, shin-etal-2020-autoprompt,schick-etal-2020-automatically}, and prototypical verbalizer~\cite{cui-etal-2022-prototypical}. For all of these experiments, we apply experimental setups similar to PET paper, but only replace the manual verbalizer with the automatic verbalizer \cite{schick-schutze-2021-exploiting}. Table~\ref{table:verbalizer} shows the average accuracy over three runs with three different seeds on different datasets with these verbalizers. From Table \ref{table:verbalizer}, the prototypical verbalizer shows better performance than other verbalizers for three (Yahoo, RTE, and MNLI) out of five datasets. The search verbalizer and soft verbalizer models perform better than the prototypical verbalizer model only on one dataset each. Since the prototypical verbalizer performs better than other verbalizers in majority of the datasets, we decided to use this as our automatic verbalizer. 
%Our results also align with the paper that proposes the prototypical verbalizer in the supervised learning setting \cite{cui-etal-2022-prototypical}.
% \begin{table}[]
% \small
% \centering
% \resizebox{\columnwidth}{!}{%
% \begin{tabular}{|l|r|rr|}\hline
% \textbf{Datasets}   & \textbf{}            & \multicolumn{2}{c|}{\textbf{SSL PET}} \\\hline  
% \textbf{}           & \textbf{\# Training} & \textbf{ProtoVerb}               & \textbf{Manual}               \\\hline  \hline  
% \multicolumn{4}{|c|}{Topic Classification}  \\\hline
% AG’s News  & 1        & 79.97    & 80.7  \\
% AG’s News  & 5        & 87.3     & 87.8  \\
% AG’s  News & 10       & 88.7     & 88.8  \\
% AG’s News  & 20       & 89.2     & 89.2  \\\hline  
% Yahoo      & 1        & 62.0     & 62.3  \\
% Yahoo      & 5        & 67.8     & 68.0  \\
% Yahoo      & 10       & 70.0     & 69.5  \\
% Yahoo      & 20       & 70.9     & 70.7  \\\hline  
% \textbf{TC Avg} &     & 77.0     & 77.1  \\\hline\hline  
% \multicolumn{4}{|c|}{Natural Language Inference}\\\hline
% MNLI    & 1       & 44.2     & 44.8  \\
% MNLI    & 5       & 55.3     & 55.2  \\
% MNLI    & 10      & 62.3     & 60.5  \\
% MNLI    & 20      & 69.6     & 68.6  \\\hline
% CB      & 32      & 85.7     & 86.9  \\\hline
% RTE     & 32      & 52.8     & 58.8  \\\hline
% \textbf{NLI Avg}& & 61.7     & 62.5  \\\hline\hline
% \textbf{Overall Avg} & & 70.4 & 70.8 \\\hline
% \end{tabular}
% }
% \caption{\label{table:protoverbalizer} Few-shot experiment results on different datasets by replacing manual verbalizers with Prototypical verbalizers in PET SSL setup.}
% \end{table}

\begin{table*}[]

\centering
\resizebox{\textwidth}{!}{%
\begin{tabular}{|l|c|ccccc|cc|}
\hline
\multicolumn{2}|{l}{}      & \multicolumn{5}{|c|}{\textbf{Semi Supervised Learning PET}} &  \multicolumn{2}{|c|}{\textbf{Supervised} } \\\hline
\textbf{Dataset}     & \textbf{\# Training} & \textbf{Demo+Soft} & \textbf{Vary Soft}&\textbf{Fixed Soft} &\textbf{Protoverb}& \textbf{Manual} & \textbf{Fine-Tune}&\textbf{Demo+Soft}\\\hline\hline
\multicolumn{9}{|c|}{\textbf{Topic Classification}}                     \\\hline
AG’s   News & 1           & {\bf 83.5}  &81.3            &82.8            &80.0           & 80.7            &25.7   & 62.2\\
AG’s News   & 5           & 87.6       &{\bf 88.0}       &87.3            & 87.3            & 87.8           &32.6   & 84.9\\
AG’s   News & 10          & 88.3       &88.3             &86.5            & 88.7            & \textbf{88.8}  &58.3   & 87.2\\
AG’s News   & 20          & 88.8       &{\bf 89.3}       &88.9            & 89.2            & 89.2           &86.1   & 88.0\\\hline
Yahoo       & 1           & 61.1       &{\bf 62.9}       &59.6            & 62.0            & 62.3           & 10.7  &  55.6\\
Yahoo       & 5           & 67.4       & 67.9            &67.1            & 67.8            &\textbf{68.0}   & 12.1  & 65.2\\
Yahoo       & 10          & 68.9       & 69.5            &69.1            &\textbf{70.0}    & 69.5           & 37.8  & 67.0\\
Yahoo       & 20          & 70.7       & {\bf 71.0}            &70.4            &70.9   & 70.7           & 66.7  &  66.5\\\hline\hline
\textbf{TC Avg} &    -     & 77.0       &\textbf{77.3}    &76.5            &77.0             & 77.1           & 41.2      &72.1  \\\hline\hline
\multicolumn{9}{|c|}{\textbf{Natural Language Inference}}               \\\hline  
MNLI     & 1              & 36.1       &51.7            &\textbf{52.7}   &44.2             & 44.8   &    34.3  & 35.1 \\
MNLI     & 5              & 51.2       &{\bf 58.1}     &57.7            &55.3             & 55.2   &     33.5 & 46.9 \\
MNLI     & 10             & 60.4       &57.8            &58.4            &\textbf{62.3}    & 60.5   &    34.3  & 54.4 \\
MNLI     & 20             & 64.0       &64.7            &60.5            &\textbf{69.6}   & 68.6    &    35.0  & 41.9 \\\hline
CB       & 32             &{\bf 88.7} &88.1            &88.7   &85.7             & 86.9   &   60.7    &  87.6\\\hline
RTE      & 32             &{\bf 70.4} &62.5            &62.6            &52.8             & 58.8   &   48.1    &  67.4\\\hline\hline
\textbf{NLI Avg}&   -  & \textbf{70.7}   & 69.6  & 69.5 & 65.5 & 67.7   & 47.7     & 66.5\\\hline\hline

\textbf{Overall Avg } & -  & \textbf{73.2}  & 72.6 &72.3  &70.1 & 71.4  &  45.1 & 68.7\\\hline
\end{tabular}
}
\caption{\label{table:results}   Few-shot experiment results (average accuracy) on different datasets with our proposed methods in PET SSL setup. For CB and RTE, we use $32$ training examples, whereas for other datasets we use $\{1, 5, 10, 20\}$ randomly selected examples per class for few-shot learning experiments. The best performance is marked in bold. Note that to report the average results for NLI task, we first average over the MNLI results under different few-shot settings, and then average over the three NLI datasets to give each task equal weight. The overall average results are computed following a similar approach, giving each dataset an equal weight.}
\end{table*}

\subsection{Comparison with Manual PET}
With the prototypical verbalizer as our automatic verbalizer, we then experiment with our proposed methods for automatic prompt design. Table~\ref{table:results} shows our results on different datasets and tasks in the few-shot setting. 
%We use  ${\{1 ,5, 10, 20\}}$ number of instances per label for our few-shot experiments for most datasets. For CB and RTE, we follow previous research directions and use $32$ instances for training \cite{schick2020s}.
Table~\ref{table:results} shows that by only replacing the manual verbalizer with the prototypical verbalizer (column \textbf{Protoverb}) and keeping other aspects of the experiment the same as the PET method, we can achieve slightly lower performance ($70.1$ average accuracy) compared to Manual PET ($71.4$ average accuracy)~\cite{schick-schutze-2021-exploiting}. This shows that to eliminate human involvement in designing verbalizers, we can simply replace the manual verbalizer with the prototypical verbalizer with only a little performance sacrifice.

For our next set of experiments, we replace manual prompts with our proposed method, automatically creating multiple prompts. The first method (Demo+Soft Tokens PET), which adds randomly sampled demonstration examples from training data with a fixed number of trainable continuous prompt tokens with input, achieves better performance than Manual PET method. The next method (Vary Soft PET), in which we vary the number of continuous trainable tokens, also achieves better performance than Manual PET method. For topic classification tasks, under multiple few-shot settings, the average accuracy of Demo+Soft and Vary Soft PET are $77.0$ and $77.3$, respectively, while the average accuracy of Manual PET method is $77.1$. Similarly, for NLI datasets under different few-shot settings, the average accuracy of our Vary Soft PET method is $69.6$ and Demo+Soft Tokens PET method is $70.7$. Both of these results are better than Manual PET method ($67.7$). Furthermore, across all these datasets, Demo+Soft Tokens PET and Vary Soft PET achieve an average performance of $73.2$ and $72.6$, respectively. These results are better than Manual PET ($71.4$) method. This experiment shows that it is possible to completely eliminate human involvement and expertise in designing prompts and verbalizers for the SSL pipeline with even better performance.

We also observe that for the case of one-shot experiments with MNLI dataset, Demo + Soft PET method obtains an accuracy of $36.1$, which is much worse than other prompt baseline models. This may be due to randomly sampled $[demo]$ examples, as previous studies have shown that the choice of examples in the few-shot setting can result in high-variance performance \cite{lu2021fantastically}. In future work, we can utilize sentence embeddings to make intelligent decisions while selecting demonstration examples.

% Compared with Vary Soft PET method, the first Demo+Soft PET does not have such a performance improvement. The reason may originate in randomly sampled $[demo]$ and previous studies show that the choice of examples in the few-shot setting can result in high-variance performance \cite{lu2021fantastically}. In future work, we may utilize the sentence embeddings to select the closest demonstration examples.
%}
 
\subsection{Ablation Study}
\subsubsection{Impact of Semi-supervised Learning}
%supervised experiments
We compare our proposed methods with supervised learning methods: fine-tuning and prompt-based tuning methods (Demo+Soft in SL). All semi-supervised learning methods perform significantly better than supervised learning methods. Traditional fine-tuning methods perform the worst ($45.1$ average accuracy) on different datasets and tasks. Demo+Soft in SL method is similar to our proposed Demo+Soft Tokens PET method but does not make use of unlabeled data. Demo+Soft in SL performs better than the fine-tuning method and achieves an average accuracy of $68.7$ on multiple datasets and tasks in different few-shot settings. Both of the supervised learning methods perform worse than any SSL prompting model, indicating the necessity of the SSL pipeline in NLU tasks. 
%The prompt-based tuning method is similar to our proposed SSL method, which also adds demonstration and a certain number of  trainable continuous prompt tokens with prototypical verbalizer, but does not make use of unlabeled data.
\subsubsection{Impact of Diversity in the Prompts}
In order to understand the effect of introducing diversity through multiple prompts in SSL, we devise another experiment, where we use the SSL setup but use only \textbf{one} prompt labeler model (not adding a demonstration example but using trainable soft tokens) to label unlabeled data. We name this method as Fixed Soft Tokens PET. Table~\ref{table:results} shows that in most comparisons (13/14), our proposed Vary Soft PET or Demo+Soft PET method achieves better performance. When comparing with the Fixed Soft PET, our proposed Demo+Soft PET  shows an improvement of average accuracy from $72.3$ to $73.2$ ($p < 0.05$ by paired $t$ test) \cite{hsu2014paired}. Moreover, both Demo+Soft and Vary Soft PET methods obtain better average performance than the Fixed Soft Tokens PET in NLI and topic classification tasks. These results show the importance of diversity introduced by multiple prompt labeler models.
\section{Related Work}

\subsection{Language Model Prompting}
\citet{cui-etal-2021-template} authors fine-tuned the pre-trained generative language model, BART, with a predefined template (${candidate\_span}$ is a ${entity\_type}$ entity) for NER classification. 
% This method requires scanning over each candidate span of the input sequence to obtain the NER label during inference time. They showed competitive results in a rich resource NER setting (CONLL03 dataset) and significantly better performance in a low resource NER setting (cross-domain and few-shot NER benchmark datasets). 
\citet{wang2021entailment} proposed Entailment as Few-shot Learner (EFT) method, which transforms classification tasks into natural language textual entailment tasks and then fine-tunes the LM. The transformation also makes it easy to leverage unsupervised contrastive data augmentation methods to add pairwise examples to the limited annotated data. This setting further showed an average 2.7\% improvement in 15 different NLP tasks. In addition to using the prompts for supervised learning, PET is the SoTA method to adapt the manual prompts along with semi-supervised learning to obtain strong performance across multiple NLU tasks. \cite{schick-schutze-2021-exploiting}. 

\subsection{Automatic Prompts and Verbalizers}
\citet{shin-etal-2020-autoprompt} used a gradient-guided search to find the discrete tokens for prompts based on task accuracy, initialize tokens, and then fine-tune the LM.  For automatic label token selection, they first train a logistic regression classifier from the contextualized embedding of the MASK token and then predict the score from MLM’s output word embeddings. They select the top-k highest scoring words for each label. They showed better performance over manual prompting methods for sentiment classification and textual entailment tasks. Similarly, instead of using a gradient-guided search for prompt tokens,  \citet{li-liang-2021-prefix} and \citet{lester-etal-2021-power} attached prefix vectors and learned the embeddings for prefix vectors by keeping the LM model parameters frozen. 
% \citet{li-liang-2021-prefix}  prepended prefix vectors to input layer and each layer of the encoder stack, whereas  \citet{lester-etal-2021-power} does not prepend prefix vectors to each network layer, resulting in fewer parameters than \citet{li-liang-2021-prefix}.  
% \citet{lester-etal-2021-power} also shows the effectiveness of prompt emsembling over the single prompt by training multiple prompts for the same task. 
\citet{liu2021gpt} proposed P-tuning, which replaces the input embeddings of pre-trained language models with its differentiable output embeddings, using the pattern based on human design. \citet{P2} optimized and adapted the Prefix Tuning model for NLU. \citet{vu2021spot} proposed to learn soft prompt embeddings from one or more source tasks and then transfer them to initialize the prompts for the target task. In addition, they also proposed an efficient retrieval approach to find task embeddings and predict the most transfarable source tasks for a given novel target task.

Several automatic verbalizers, such as search-based verbalizers, soft verbalizers, and prototypical verbalizers, have been proposed to automate the design of the verbalizer mapping function. Search-based verbalizers aim to find the appropriate tokens to replace human selection \cite{schick-etal-2020-automatically, shin2020autoprompt, gao2020making}. Both soft verbalizers and prototypical verbalizers learn trainable class or prototyope embeddings during the training process \cite{cui-etal-2022-prototypical, zhang2021differentiable, hambardzumyan-etal-2021-warp}.

\citet{mahabadi2022perfect}  proposed a prompt-free method (PERFECT) to train the language model, which does not rely on manual commands and verbalizers. PERFECT reported performance similar to that of PET~\cite{schick-schutze-2021-exploiting} in a few-shot setting. However, they used a supervised learning setup and compared their results with the single labeler model with one prompt rather than the results from the final classifier. Here, we use a similar SSL setting to~\citet{schick-schutze-2021-exploiting} and report the results of the final classifier.

\section{Conclusions}
In this paper, we are able to successfully use automatic prompts and verbalizers in semi-supervised learning settings. We show that our proposed automatic prompt generation methods with prototypical verbalizer can eliminate human engineering in prompt-based SSL setup and achieve similar or better performance than the SoTA Manual PET method. Our methods have the added advantage of being scalable with multiple tasks and datasets. We also empirically verify the power of semi-supervised learning methods, which take advantage of large amounts of unlabeled data, over supervised methods. 

In the next steps, we plan to investigate whether we would be able to achieve similar performance by freezing PLMs' parameters and only tuning verbalizer and prompt parameters. This setup will save a tremendous amount of space by making it easy to share and reuse PLMs. Moreover, we plan to explore ways to combine the two proposed methods Demo+Soft PET and Vary Soft PET, which would take advantage of both methods.

\section{Limitations}
% We conduct the experiment with only one type of PLM: RoBERTa-Large model, and we can also integrate our Vary Soft PET and Demo+Soft PET with other effective language models such as ALBERT and T5 \cite{lan2019albert, raffel2020exploring}. Among various NLU tasks, the proposed methods show effectiveness in two tasks: text classification and natural language inference. We can generalize our framework to other NLU tasks, such as word-sense disambiguation and paraphrase detection, in the further study.
Although we experiment with multiple NLU tasks and datasets, these datasets are only in the English language. Prompt-based learning relies on large language models, which have acquired knowledge through pre-training on huge corpora. With low-resource languages, it might be difficult to get PLMs trained on a huge corpus, which might make it hard to reproduce performance similar to the English corpus. The fine-tuning and inference of PLM requires multiple large GPUs, which might not be accessible to everyone. 
%Also, in this work we use multiple prompt-based learning labeler models, each with its own PLM and weights. This demands large storage and increase latency when we want to productionize the system. 

%We will also find a way to combine the two proposed methods Demo+Soft PET and Vary Soft PET into a unified method, which takes the diversity of both demonstration examples and the varying number of soft tokens.

\section*{Acknowledgments}
We would like to thank  the anonymous reviewers as well
as Wei Ai, Paiheng Xu, Akram Almatarky, Jangwon Kim, Morteza Ziyadi, and  Giannis Karamanolakis for
reviewing the paper and for providing helpful comments and suggestions. 

%We would also like to thank all the anonymous reviewers for the helpful feedback and suggestions on this work. 

% This work was supported in part by Amazon Alexa AI.
% The acknowledgments should go immediately before the references. Do not number the acknowledgments section.
% \textbf{Do not include this section when submitting your paper for review.}
\bibliographystyle{acl_natbib}
% \bibliography{anthology,custom}

\appendix
\section{Prompts and Verbalizers}
\label{templates}
\subsection{Manual Prompts and Manual Verbalizers}

We use the same  manual prompts and manual verbalizers for our baseline experiment as used by  \citet{schick-schutze-2021-exploiting, schick2020s}.

 \textbf{AG's News} is a news topic classification dataset with four classes. We use the manual verbalizer that maps class 1-4 to ``World”, ``Sports”, ``Business” and ``Technology”. For the input sentence $x=(a, b)$, where $a$ is the news headline and $b$ is the body of the news text, we use the manual prompts below:
\begin{align*}
    & \mathcal{P}_1(x) = \text{[MASK] : } [a] \; [b] \\
    & \mathcal{P}_2(x) = \text{[MASK] - } [a] \; [b] \\
    & \mathcal{P}_3(x) = [a] \; (\text{[MASK]}) \; [b] \\
    & \mathcal{P}_4(x) = [a] \; [b] \; (\text{[MASK]}) \\
    & \mathcal{P}_5(x) = \text{[MASK] News: } [a] \; [b] \\
    & \mathcal{P}_6(x) = \text{Category : [MASK]} \; [a] \; [b]
\end{align*}

\textbf{Yahoo Questions} is another dataset for topic classification with ten classes. We use the same manual  prompts as AG's News, but define the manual verbalizer for the Yahoo dataset, which maps the classes 1-10 to ``Society”,
``Science”, ``Health”, ``Education”, ``Computer”,
``Sports”, ``Business”, ``Entertainment”, ``Relationship” and ``Politics”.

\textbf{MNLI} is the dataset for textual entailment tasks, consisting of text pairs $x=(a, b)$. We define two manual verbalizer pairs $v_1$ and $v_2$. $v_1$ verbalizer maps class 0-2 to ``Wrong", ``Right" and ``Maybe". $v_2$ verbalizer maps class 0-2 to ``No", ``Yes", ``Maybe".  We use the following manual prompts:
\begin{align*}
    & \mathcal{P}_1(x) = ``[a]" \; \text{? \(\vert \vert\) [MASK],} \;  ``[b]" \\
    & \mathcal{P}_2(x) = [a] \; \text{? \(\vert \vert\) [MASK],} \;  [b] 
\end{align*}

\textbf{RTE} and \textbf{CB} are  datasets for textual entailment tasks. We use $v_1$ as the manual verbalizer similar to MNLI task. We use the following manual prompts:
\begin{align*}
    & \mathcal{P}_1(x) = ``[a]" \; \text{? \(\vert \vert\) [MASK],} \;  ``[b]" \\
    & \mathcal{P}_2(x) = [a] \; \text{? \(\vert \vert\) [MASK],} \;  [b] \\
    & \mathcal{P}_3(x) = [a] \; \text{? \(\vert \vert\) [MASK].} \;  [b] \\
    & \mathcal{P}_4(x) =  ``[a]" \; \text{? \(\vert \vert\) [MASK].} \;  ``[b]" 
\end{align*}

% These manual prompts and verbalizers  are the same as defined by \citet{schick2020s}.

\subsection{Continuous Prompts}
\textcolor{black}{
For our proposed models: \textbf{Demo+Soft} and \textbf{Vary Soft} models, we apply continuous prompts and automatic verbalizers to ensure that the prompt-tuning SSL method can be scaled across multiple datasets. From  previous works, we find that few anchor tokens help to improve the performance of NLU tasks \cite{liu2021gpt}, so we design two different continuous prompts dependant on the nature of NLU tasks.  
For the continuous prompt for AG's News and Yahhoo Questions (text classification task), our design is:
\begin{equation*}
   \mathcal{P}(x) = [a] \; [b] \; \text{Category: } \; [p_0,p_1,\ldots,p_{n}] \; [\text{MASK}]  \\ 
\end{equation*}
For continuous prompt for MNLI, CB and RTE (NLI tasks), our design is:
\begin{equation*}
   \mathcal{P}(x) = [a] \; [b] \; \text{?} \; [p_0,p_1,\ldots,p_{n}] \; \text{answer : [MASK]} \\ 
\end{equation*}
The construction of continuous prompts also follow the design of the P-tuning paper \cite{liu2021gpt}. Rather than designing multiple manual prompts for different datasets, we can use our proposed methods to automate this process. This reduces  human efforts and costs when we scale across multiple datasets and tasks.
}
% Compared with designing multiple manual prompts for different datasets, only one design of continuous prompts for one task and no need to design verbalizers make the human efforts not as the bottleneck for prompt-tuning SSL method to generalize on scalabel datasets. 
\end{document}